\documentclass{article}

 \usepackage[preprint]{neurips_2025}
\usepackage{amsmath,amsfonts,amsthm,amsthm,amssymb}
\usepackage{bm,bbm}
\newtheorem{theorem}{Theorem}
\newtheorem{lemma}{Lemma}

\usepackage{varwidth}

\usepackage{hyperref}
\usepackage{cleveref}
\usepackage{mathtools}
\usepackage{algpseudocode}
\usepackage{algorithm}
\usepackage{mdframed}

\usepackage{multicol}
\usepackage{multirow}
\usepackage{setspace}

\usepackage{colortbl}

\usepackage[svgnames]{xcolor}
\usepackage{framed}

\usepackage{tikz,ifthen}
\usetikzlibrary{positioning}
\usetikzlibrary{shapes}
\usetikzlibrary{arrows}
\usetikzlibrary{fit}
\usetikzlibrary{calc}
\usetikzlibrary{shapes.misc}

\newcommand*\openquote{\makebox(25,0){\scalebox{3}{``}}}
\newcommand*\closequote{\makebox(25,0){\scalebox{3}{''}}}
\colorlet{shadecolor}{Azure}

\makeatletter
\newif\if@right
\def\shadequote{\@righttrue\shadequote@i}
\def\shadequote@i{\begin{snugshade}\openquote}
\def\endshadequote{%
  \if@right\hfill\fi\closequote\end{snugshade}}
\@namedef{shadequote*}{\@rightfalse\shadequote@i}
\@namedef{endshadequote*}{\endshadequote}
\makeatother

\crefname{defi}{defn.}{defns.}

\def\1{\bm{1}}

\def\vtheta{{\bm{\theta}}}

\def\vu{{\bm{u}}}

\def\vx{{\bm{x}}}

\def\mI{{\bm{I}}}

\DeclareMathAlphabet{\mathsfit}{\encodingdefault}{\sfdefault}{m}{sl}
\SetMathAlphabet{\mathsfit}{bold}{\encodingdefault}{\sfdefault}{bx}{n}

  \usepackage{listings}
  \usepackage{xcolor} % For defining custom colors

  \definecolor{codegreen}{rgb}{0,0.6,0}
  \definecolor{codegray}{rgb}{0.5,0.5,0.5}
  \definecolor{codepurple}{rgb}{0.58,0,0.82}
  \definecolor{backcolour}{rgb}{0.95,0.95,0.92}

  \lstdefinestyle{mystyle}{
      backgroundcolor=\color{backcolour},
      commentstyle=\color{codegreen},
      keywordstyle=\color{magenta},
      numberstyle=\tiny\color{codegray},
      stringstyle=\color{codepurple},
      basicstyle=\ttfamily\footnotesize,
      breakatwhitespace=false,
      breaklines=true,
      captionpos=b,
      keepspaces=true,
      numbers=left,
      numbersep=5pt,
      showspaces=false,
      showstringspaces=false,
      showtabs=false,
      tabsize=2,
      language=Python, % Specify the language
  }

      \crefname{lstlisting}{listing}{listings}
    \Crefname{lstlisting}{Listing}{Listings}

\usepackage[utf8]{inputenc} % allow utf-8 input
\usepackage[T1]{fontenc}    % use 8-bit T1 fonts
\usepackage{hyperref}       % hyperlinks
\usepackage{url}            % simple URL typesetting
\usepackage{booktabs}       % professional-quality tables
\usepackage{amsfonts}       % blackboard math symbols
\usepackage{nicefrac}       % compact symbols for 1/2, etc.
\usepackage{microtype}      % microtypography
\usepackage{xcolor}         % colors

\title{Gaussian Embeddings: How JEPAs\\Secretly Learn Your Data Density}

\author{%
  Randall Balestriero \\
  Meta-FAIR \& Brown University\\
  \texttt{rbalestr@brown.edu} \\
  \And
  Nicolas Ballas \\
  Meta-FAIR \\
  \And
  Mike Rabbat \\
  Meta-FAIR \\
  \And
  Yann LeCun \\
  Meta-FAIR \& NYU
}

\begin{document}

\maketitle

\begin{abstract}
Joint Embedding Predictive Architectures (JEPAs) learn representations able to solve numerous downstream tasks out-of-the-box. JEPAs combine two objectives: (i) a latent-space prediction term, i.e., the representation of a slightly perturbed sample must be predictable from the original sample's representation, and (ii) an anti-collapse term, i.e., not all samples should have the same representation. While (ii) is often considered as an obvious remedy to representation collapse, we uncover that JEPAs' anti-collapse term does much more--it provably estimates the data density. In short, any successfully trained JEPA can be used to get sample probabilities, e.g., for data curation, outlier detection, or simply for density estimation. Our theoretical finding is agnostic of the dataset and architecture used--in any case one can compute the learned probabilities of sample $\vx$ efficiently and in closed-form using the model's Jacobian matrix at $\vx$. Our findings are empirically validated across datasets (synthetic, controlled, and Imagenet) and across different Self Supervised Learning methods falling under the JEPA family (I-JEPA and DINOv2) and on multimodal models, such as MetaCLIP. We denote the method extracting the JEPA learned density as {\bf JEPA-SCORE}.
\end{abstract}

\begin{figure}[h]
    \centering
    \begin{minipage}{0.02\linewidth}
    \rotatebox{90}{\small MetaCLIP\;\;IJEPA-22k\;\;IJEPA-1k\;\;DINOv2\;\;\;\;\;\;}
    \end{minipage}
    \begin{minipage}{0.97\linewidth}
    \centering
    \begin{minipage}{0.49\linewidth}
    \centering\color{blue}
        {\bf low probability}\\[-0.6em]
        \makebox[0pt][c]{}\rule{\linewidth}{1.6pt}
    \end{minipage}
    \begin{minipage}{0.49\linewidth}
    \centering\color{red}
    {\bf high probability}\\[-0.6em]
        \makebox[0pt][c]{}\rule{\linewidth}{1.6pt}
    \end{minipage}
    \includegraphics[width=\linewidth]{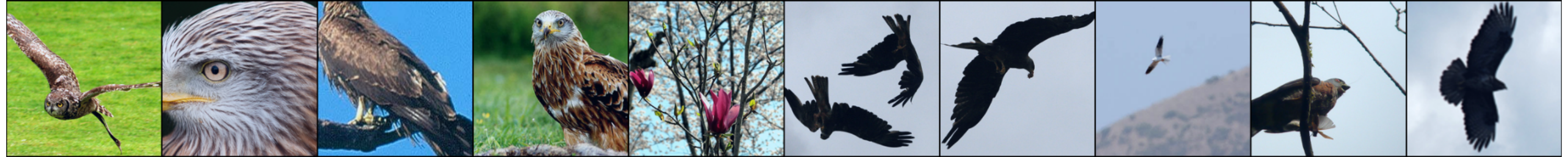}\\
    \includegraphics[width=\linewidth]{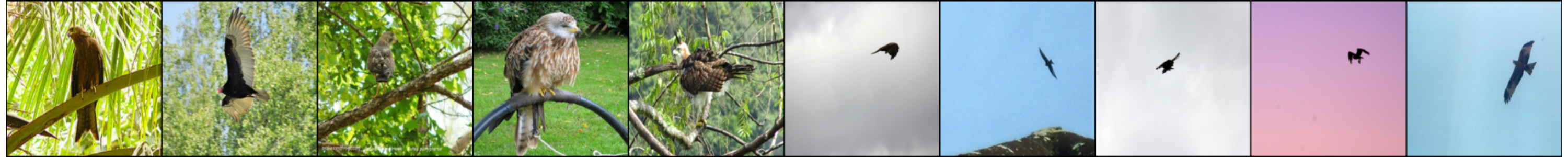}\\
    \includegraphics[width=\linewidth]{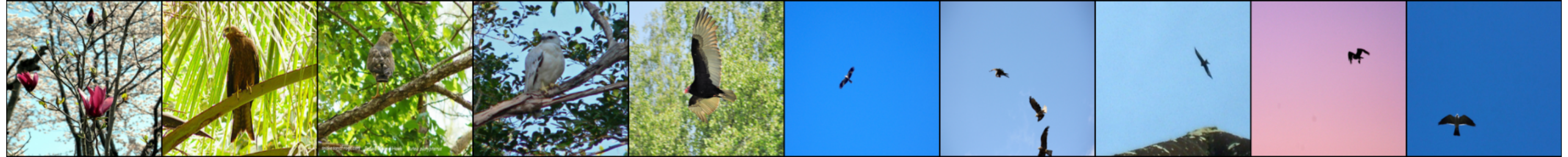}\\
    \includegraphics[width=\linewidth]{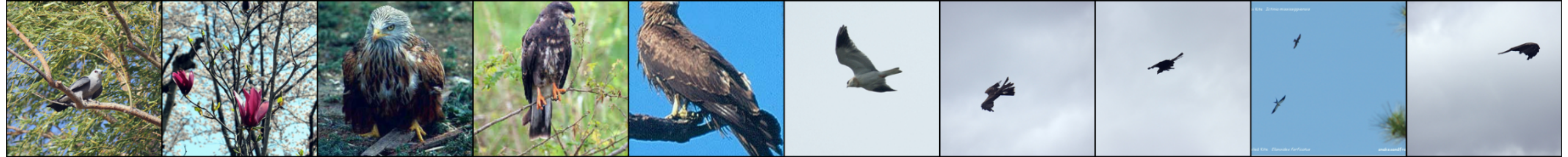}
    \end{minipage}
    \caption{\small Depiction of the 5 least ({\bf left}) and 5 most ({\bf right}) likely samples of class 21 from Imagenet as per {\bf JEPA-SCORE}--JEPAs' implicit density estimator learned during pretraining. Two striking observations: (i) across all JEPAs ({\bf rows}) the type of samples with low and high probabilities are alike, and (ii) the same samples (amongst 1,000) are found at those extrema. Random samples from that class are provided in \cref{fig:samples_1}}
    \label{fig:placeholder}
\end{figure}
\vspace{-0.2cm}
\section{Introduction}
\vspace{-0.2cm}
The training procedure of foundation models---Deep Networks (DNs) $f_{\vtheta}$ able to solve many tasks in zero or few-shot---can take many forms and is at the center of Self Supervised Learning research \cite{balestriero2023cookbook}. Over the years, one principle has emerged as central to all current state-of-the-art methods: encouraging $f_{\vtheta}(X)$ to be maximum Entropy given i.i.d pretraining samples $X$ with density $p_{X}$ \cite{wang2020understanding,hjelm2018learning}. Because the differential Entropy is difficult to estimate in high-dimensional spaces, and $f_{\vtheta}(X)$ often contains thousands of dimensions, one solution is to maximize a lower bound by using a decoder and learning to reconstruct $X$ from $f(X)$ \cite{vincent2008extracting}. Because this approach comes with known limitations \cite{balestriero2024learning}, more and more foundation models are trained with Joint-Embedding Predictive Architectures (JEPAs) \cite{lecun2022path} that directly encourage $f_{\vtheta}(X)$ to be Gaussian.  In fact, the Gaussian distribution is the one of maximum differential Entropy under covariance constraint, leading to $f_{\vtheta}(X)$ producing {\em Gaussian Embeddings} (GE).

JEPAs can take many forms by employing numerous implicit and explicit regularizers \cite{srinath2023implicit,littwin2024jepa}. Today's JEPAs mostly take three forms: (i) moment-matching objectives (VICReg \cite{bardes2021vicreg}, W-MSE \cite{ermolov2021whitening}), (ii) non-parametric estimators (SimCLR \cite{chen2020simple}, MoCo \cite{he2020momentum}, CLIP \cite{radford2021learning}), and (iii) implicit teacher-student methods (DINO \cite{caron2021emerging}, I-JEPA \cite{assran2023self}). While JEPAs produce state-of-the-art representations, they are currently seen as disconnected from generative models whose goal is to estimate $p_X$. In fact, the absence of generative modeling is a praised benefit of JEPAs. But one may wonder\dots
\begin{center}
{\em Can the density of $f(X)$ be specified without $f$ learning about $p_X$?}
\end{center}
That question will be at the core of our study, and the answer is no: producing Gaussian Embeddings can only happen if $f_{\vtheta}$ learns the underlying data density $p_{X}$. But {\bf JEPAs estimate $p_{X}$ in a highly non standard way, free of input space reconstruction, and free of a parametric model for $p_X$}. One question remains\dots
\begin{center}
{\em Is there any further benefit of not only specifying a density for $f_{\vtheta}(X)$\\but using the eponymous Gaussian density?}    
\end{center}
At it turns out, this choice guarantees that the estimator for $p_X$ implicitly learned during JEPA training can easily be extracted from the final trained model $f_{\vtheta}$--an estimator we call the {\bf JEPA-SCORE} (\cref{eq:JEPA_SCORE}). Our findings not only open new avenues in using {\bf JEPA-SCORE} for outlier detection or data curation, but also shaken the Self Supervised Learning paradigm by showing how {\bf non parametric density estimation in high dimension is now amenable through JEPAs}. Our theory and its corresponding controlled experiments are provided in \cref{sec:uniform,sec:density}, and experiments with state-of-the-art models such as DINOv2, MetaCLIP and I-JEPA are provided in \cref{sec:implementation}. {\bf JEPA-SCORE}'s implementation only takes a few lines of code and is provided in~\cref{code}.

\vspace{-0.2cm}
\section{JEPA-SCORE: the Data Density Implicitly Learned by JEPAs}
\vspace{-0.2cm}

We now derive our main result stating that in order to minimize the JEPA objective, a DN must learn the data density. We start with some preliminary results in \cref{sec:uniform}, we formalize our general finding in \cref{sec:density}, culminating in the JEPA result of \cref{sec:implementation,thm:JEPA}. An efficient implementation is also provided in \cref{sec:implementation}.

\vspace{-0.2cm}
\subsection{Preliminaries: Gaussian Embeddings Are Uniform on the Hypersphere}
\label{sec:uniform}
\vspace{-0.2cm}

Our derivations will rely on a simple observations widely known in high-dimensional statistics: $K$-dimensional standard Gaussians, appropriately normalized, converge to being Uniform on the hypersphere. Let's denote the $K$-dimensional standard Gaussian random variable by $Z$, and the normalized version by $X\triangleq \frac{Z}{\sqrt{K}}$ with density $f_{\mathcal{N}(0,\mI/K)}$. Let's also denote the Uniform distribution on the $K$-dimensional hypersphere surface by $f_{\mathcal{U}(\mathbb{S}(0,R,K))}$ with radius $R>0$.

\vspace{-0.1cm}
\begin{lemma}
    As $K$ grows, $X$ quickly concentrates around the hypersphere of radius $1$, converging to a Uniform density over the hypersphere surface. (Proof in \cref{proof:uniform}.)
    \label{thm:uniform}
\end{lemma}
\vspace{-0.1cm}

% As the proof is short and offers practical insights for the rest of our discussions, we provide it below.

\Cref{thm:uniform} provides an interesting geometric fact which we turn into a practical result for SSL in the following \cref{sec:density}, where we demonstrate how learning to produce Gaussian embeddings equates with learning the Energy function of the training data.

\vspace{-0.2cm}
\subsection{Producing Gaussian Embeddings Equates Learning an Energy Function}
\label{sec:density}
\vspace{-0.2cm}

This section builds upon \cref{thm:uniform} to demonstrate how learning to produce Gaussian embeddings implies learning about the data density.

Consider two densities, one on the input domain ($p_X$) and one on the output domain ($p_{f(X)}$). For $p_{f(X)}$ to have a particular form, e.g., $\mathcal{N}(0,\mI/K)$, $f$ must learn something about $p_X$. To see that, we will have leverage the eponymous change of variable formula expressing the embedding density $p_{f_{\vtheta}}$ as a function of the data density and the DN's Jacobian matrix:
% provide intuitions, let's first consider a simplified setup where $f$ is a bijection from ${\rm supp}(p_X)$ to $\mathbb{R}^{K}$, and let's consider a small region $\mathcal{R}$ of the input space--small enough for a first-order Taylor expansion of $f$ to be a good approximation. We have that $F(X) \approx \mA(\mathcal{R})X+\vb(\mathcal{R})$ and thus the output density is
% \begin{align}
% p_{f(X)|X\in\mathcal{R}}(f(\vx))=\frac{p_X(\vx)}{\prod_{k=1}^{rank(\mA(\mathcal{R}))}\sigma_k(\mA(\mathcal{R}))}.\label{eq:CPA}
% \end{align}
% Hence, it is direct to see that the model's parameter must adjust $\mA(\mathcal{R})$ to align with $p_X(\vx)$ in order for $p_{f(X)|X\in\mathcal{R}}(f(\vx))$ to match the target embedding density. The few assumptions we required to obtain \cref{eq:CPA} can be lifted, leading to a the more general embedding density for any nonlinear mapping $f$
\newcommand{\rank}{\operatorname{rank}}
\begin{equation}
    p_{f(X)}(f(\vx))=\int_{\{\vu \in \mathbb{R}^{D} | f(\vu)=f(\vx)\}} \frac{p_{X}(x)}{\prod_{k=1}^{\rank(J_{f}(\vx))}\sigma_k(J_{f}(\vx))} \mathrm{d} \mathcal{H}^{r}(\vx),\label{eq:general}
\end{equation}
where $\mathcal{H}^{r}$ denotes $r$-dimensional Hausdorff measure, with $r\triangleq \dim (\{\vu \in \mathbb{R}^{D} | f(\vu)=f(\vx)\})$ being the dimension of the level set of $f$ at $\vx$. We note that \cref{eq:general} does not require $f$ to be bijective, which will be crucial for our JEPA result in \cref{sec:implementation}; for details see \cite{krantz2008geometric,cvitkovic2019minimal}. Combining \cref{eq:general} and \cref{thm:uniform} leads us to the following result.

% For example 
% The goal of this section is to demonstrate how learning to produce Gaussian embeddings implies learning the Energy function of the data as part of the DN.

\begin{figure}[t!]
    \centering
    \begin{minipage}{0.55\linewidth}
    \includegraphics[width=\linewidth]{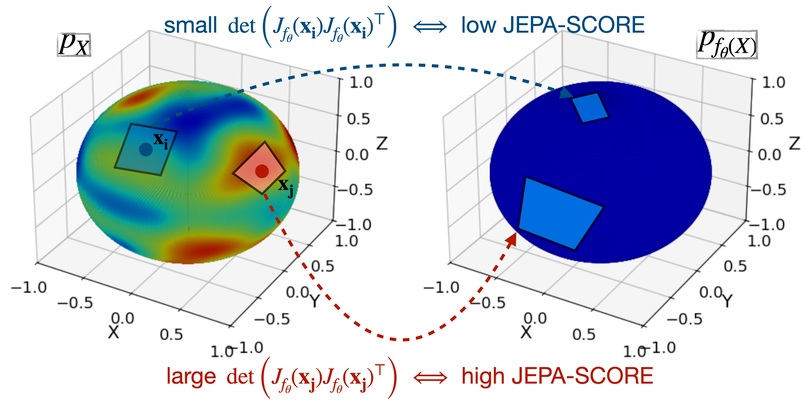}
    \end{minipage}
    \begin{minipage}{0.44\linewidth}
    \begin{tabular}{lrrrr}
\toprule
 & 512 & 1024 & 2048 & 4096 \\
dim &  &  &  &  \\
\midrule
64 & 0.64 & 0.69 & 0.84 & 0.90 \\
128 & 0.75 & 0.85 & 0.90 & 0.94 \\
256 & 0.82 & 0.83 & 0.69 & 0.76 \\
512 & 0.72 & 0.75 & 0.84 & 0.88 \\
\bottomrule
\end{tabular}
\end{minipage}
\vspace{-0.4cm}
\includegraphics[width=\linewidth]{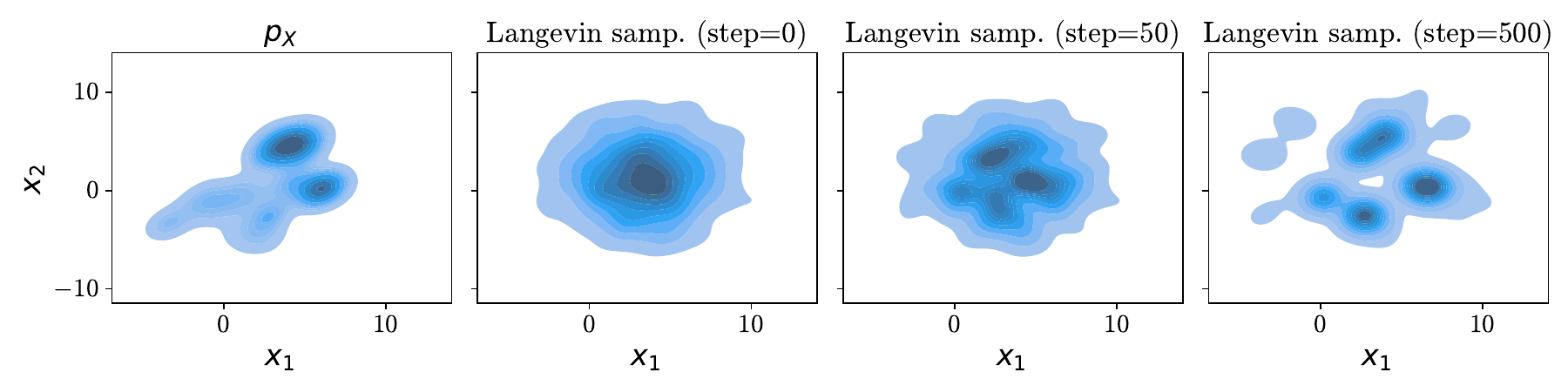}\vspace{-0.4cm}
    \caption{\small {\bf Top left:} Visual illustration of {\bf JEPA-SCORE}--the DN $f_{\vtheta}$ must learn $p_{X}$ for its Jacobian matrix to expand or contract the density in order to produce a Uniform density on the hypersphere surface in its embedding space (\cref{thm:uniform,thm:general_density}). {\bf Top right:} Pearson correlation between {\bf JEPA-SCORE} and the true $log p(x)$ on a GMM data model for various input dimensions (rows) and number of samples (columns). In all cases, producing Gaussian embeddings make the backbone $f_{\vtheta}$ internalize the data density which can be easily extracted using our proposed {\bf JEPA-SCORE}. {\bf Bottom:} as {\bf JEPA-SCORE} is an approximation of the true score function, it is possible to perform Langevin sample to recover the true data distribution as shown here in two dimensions.}
    \label{fig:density_2d}
\end{figure}

% Let's denote by $K$ the smallest integer such that there exists a $(K-1)$-dimensional compact Riemannian manifold $\mathcal{M}$ with $\int_{\mathcal{M}}p_X(\vx)d\vx=1$. That is, $K$ is the ``intrinsic dimension'' of $p_X$, sufficient to cover its support. For example, if $D=2$ but $p_X$ is a one-dimensional distribution at $\vx_2=c$, then $K=1$.
\vspace{-0.1cm}
\begin{lemma}
\label{thm:general_density}
In order for $f(X)$ to be distributed as $\mathcal{N}(0,\mI/K)$ for large $K$, $f$ must learn the data density $p_X$ up to mean-preserving rescaling within each level-set $\{\vu \in \mathbb{R}^{D} | f(\vu)=f(\vx)\}$. (Proof in \cref{proof:general_density}.)
\end{lemma}
\vspace{-0.2cm}
{\bf Empirical validation.}~Before broadening \cref{thm:general_density} to JEPAs, we first provide empirical validations that learning to produce Gaussian embeddings implies learning the data density in \cref{fig:density_2d}. We show that, in fact, the data density can be recovered with high accuracy, and it is even possible to draw samples from the estimated density through Langevin dynamics.

\vspace{-0.2cm}
\subsection{JEPA-SCORE: The Data Density Learned by JEPAs}
\label{sec:implementation}
\vspace{-0.2cm}

Most  Joint-Embedding Predictive Architecture (JEPA) methods aim to achieve two goals: (i) predictive invariance, and (ii) representation diversity, as seen in the following loss function
\begin{align}
    \mathcal{L}\triangleq&  \sum_{n=1}^{N} \mathbb{E}_{(\vx_n^{(1)},\vx_n^{(2)})\sim \mathcal{G}(\vx_n)}\left[{\rm dist}\left({\rm Pred}\left({\rm Enc}\left(\vx_n^{(1)}\right)\right),{\rm Enc}\left(\vx_n^{(2)}\right)\right)\right]&&(\text{predictive invariance})\nonumber\\
    &+{\rm diversity}\left(\left({\rm Enc}\left(\vx_{n}\right)\right)_{n \in [N]}\right),&&(\text{anti-collapse}),\label{eq:JEPA}
\end{align}
where $\vx_n^{(1)},\vx_n^{(2)}$ are two generated ``views'' from the original sample through the stochastic operator $\mathcal{G}$, and $\rm dist$ is a distance function (e.g., L2). For images, $\mathcal{G}$ typically involves two different data-augmentations. 
At this point, \cref{thm:general_density} only takes into account the non-collapse term of JEPAs. But an interesting observation from \cref{thm:general_density} is that the integration occurs over the level set of the function $f_{\vtheta}$ which coincides with the JEPA's invariance term when {\rm Pred} is near identity.

{\bf Data assumption.}~To make our result as precise as possible, we focus on the case where the views are generated from stochastic transformations, with density $p_{T}$. We also denote the density of {\em generators} as $p_{\mu}$, from which the data density $p_X$ is defined as
\begin{align}
    p_{X} \triangleq p_{\mu}\otimes p_{T}.\label{eq:data}
\end{align}
In other words, each training sample is seen as some transformation of some generator. We do not impose further restrictions, e.g., that there is only one generator per class. {\em In practical setting, the generators ($p_{\mu}$) are the original training samples prior to applying any augmentation--hence estimating $p_{\mu}$ will amount to estimating the data density.}

\begin{figure}[t!]
    \centering
    \includegraphics[width=\linewidth]{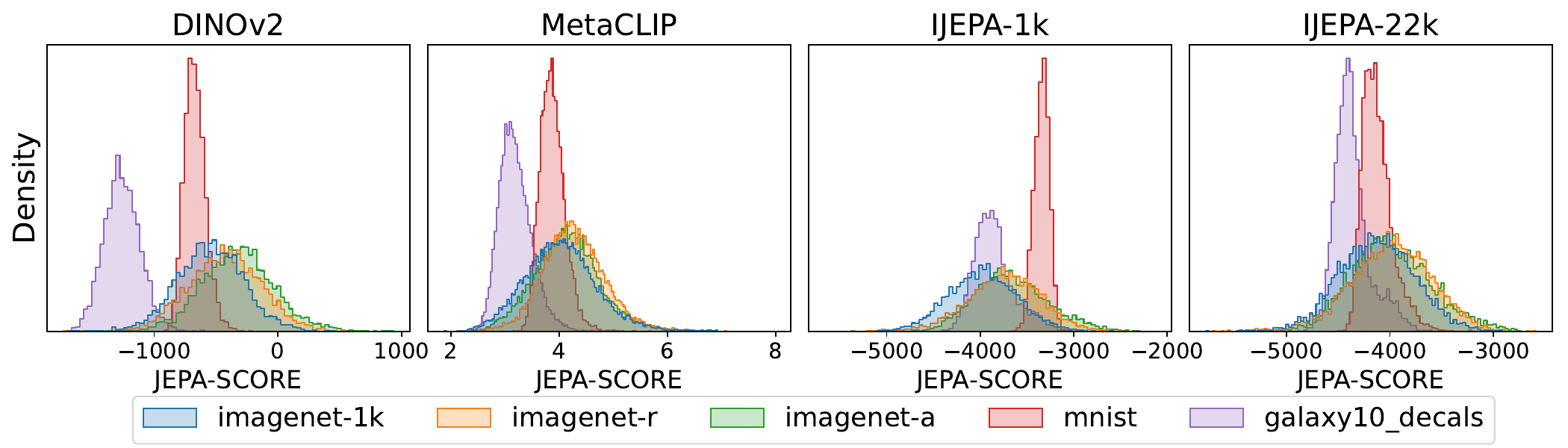}
\vspace{-0.5cm}
    \caption{\small Depiction of {\bf JEPA-SCORE} for $5,000$ samples from different datasets (imagenet-1k/a/r, MNIST and Galaxy). We clearly observe that as the pretraining dataset size increases (all models against IJEPA-1k) as MNIST and Galaxy images are seen as lower probability samples, i.e., {\em those images are less and less represented within the overall pretraining dataset}. While our score does not rely on singular vectors, we provide some examples in \cref{fig:singular_vectors}. This can be used to assess if a model is ready or not to handle particular data domains at test time for zero-shot tasks.}
    \label{fig:histograms}
\end{figure}

{\bf JEPA-SCORE.}~Combining \cref{eq:data,eq:JEPA,thm:general_density} leads to the following result proved in \cref{proof:JEPA}.

\begin{theorem}
\label{thm:JEPA}
    At optimality, JEPA embeddings estimate the data density as per
\begin{equation}
    p_{\mu}(\mu) \propto \mathbb{E}_{p_T}\left[\frac{1}{\prod_{k=1}^{\rank(J_{f}(\vx))}\sigma_k(J_{f}(\mu,T))}\right]^{-1}.\label{eq:JEPA_density}
\end{equation}
    % hereby allowing for $p_{\mu}$ to be recovered from an optimally trained JEPA model $f_{\vtheta}$.
\end{theorem}
\vspace{-0.1cm}
We define our {\bf JEPA-SCORE} for input $\vx$ as the Monte Carlo estimator of \cref{eq:JEPA_density}, for a single-sample estimate we have (in log-scale)
\begin{align}
    \text{\bf JEPA-SCORE}(\vx)\triangleq  \sum_{k=1}^{\rank(J_{f}(\vx))}\log\left(\sigma_k(J_{f}(\vx)\right),\label{eq:JEPA_SCORE}
\end{align}
which exactly recovers $p_{\mu}$ as long as the JEPA loss is minimized. We use a logarithmic scale to align with the definition of a score function. A visual illustration is provided at the top of \cref{fig:density_2d}. We empirically validate \cref{eq:JEPA_SCORE} by using pretrained JEPA models and visualizing a few Imagenet-1k classes after ordering the images based on their JEPA-SCORE in \cref{fig:placeholder,fig:samples_2,fig:samples_3}. We obtain that for bird classes, high probability samples depict flying birds while low probability ones are seated. We also conduct an additional experiment where we compute JEPA-SCORE for $5,000$ samples of different dataset (Imagenet, MNIST and Galaxy) and depict the distribution of their JEPA-SCORE in \cref{fig:histograms}. We clearly see that datasets that weren't seen during pretraining, e.g., Galaxy, have much lower JEPA-SCORE than Imagenet samples.
\\
{\bf Conclusion.}~We provided a novel connection between JEPAs and score-based methods, two families of methods thought to be unrelated until now. Controlled experiments on synthetic data confirm the validity of our theory and qualitative experiments with state-of-the-art large scale JEPAs also qualitatively validate our findings. Although this is only a first step, we hope that JEPA-SCORE will open new avenues for outlier detection and model assessment for downstream tasks.

    % %
    % Second, the mapping $f$ is continuous, and is defined on $\mathbb{R}^D$. Hence, in order for $f(\vx)$ to produce a Uniform density on the hypersphere in $\mathbb{R}^K$, $f(\vx)$ must be compact, meaning that $f$ must at least include $\mathcal{M}$ as part of its pre-image. In fact, suppose that $f$ only captured a lower dimensional sub-manifold then its output intrinsic dimension would be smaller than $K$ hereby breaking the assumption $f(\vx) \sim \mathcal{U}(0,1)$. Similarly, if $f$ captures $\mathcal{M}$ minus a ``hole'', i.e., $\mathcal{M}$ is not included in the pre-image of $f$, then its output must be non-compact too, again breaking our assumption.
    % %
    % Third, since $f$ pre-image includes $\mathcal{M}$. Given that and our assumptions, $f$ must be a diffeomorphism between $\mathcal{M}$ and the hypersphere in $R^{K}$. We now have the following known change of variable $p_{f(x)}(\vh)=p_X(\vx)\det \left( J_f(\vx)\right)^{-1}$.

\bibliography{iclr2025_conference}
\bibliographystyle{plain}

\newpage
\appendix

\section{Proofs}

\subsection{Proof of \cref{thm:uniform}}
\label{proof:uniform}

\begin{proof}

{\bf Proof 1:}~The proof consists in expressing both densities in spherical coordinates, and studying their convergence as $K$ increases. Let's first express the Uniform distribution in spherical coordinates:
\begin{align*}
    f_{\mathcal{U}(\mathbb{S}(0,R,K))}(\vx) &= \delta(\|\vx\|_2-R)\frac{\Gamma(K/2)}{2\pi^{K/2}R^{K-1}},&&\text{(Cartesian coordinates)}\\
    f_{\mathcal{U}(\mathbb{S}(0,R,K))}(r,\vtheta) &= \delta(r-R)\frac{\Gamma(K/2)}{2\pi^{K/2}}\prod_{i=1}^{K-1}\sin(\vtheta_i)^{K-i-1},&&\text{(spherical coordinates)},
\end{align*}
and let's now express the rescaled standard Gaussian density $\frac{Z}{\sqrt{K}}$ in spherical coordinates:
\begin{align*}
    f_{\mathcal{N}(0,\mI/K)}(\vx) &=\left(\frac{K}{2\pi}\right)^{K/2}e^{-\frac{K}{2}\|\vx\|_2^2},&&\text{(Cartesian coordinates)}\\
    f_{\mathcal{N}(0,\mI/K)}(r,\vtheta) &= \left(\frac{K}{2\pi}\right)^{K/2}e^{-\frac{K}{2}r^2}r^{K-1}\prod_{i=1}^{K-1}\sin(\vtheta_i)^{K-i-1}\\
     &= \frac{K^{\frac{K}{2}}}{2^{K / 2-1} \Gamma(K / 2)}  e^{-\frac{Kr^{2}}{2}}\frac{\Gamma(K/2)}{2\pi^{K/2}}r^{K-1}\hspace{-0.15cm}\prod_{i=1}^{K-1}\sin(\vtheta_i)^{K-i-1}\\
     &= \underbrace{\frac{K^{\frac{K}{2}}}{2^{K / 2-1} \Gamma(K / 2)} r^{K-1} e^{-\frac{Kr^{2}}{2}}}_{\text{scaled Chi-distribution}\overset{K\rightarrow \infty}{\rightarrow} \delta(r-1)}f_{\mathcal{U}(\mathbb{S}(0,R,K))}(r ,\vtheta),&&\text{(spherical coordinates)}.
\end{align*}
As $K$ increases, as the scaled Chi-distribution converges to a Dirac function at $1$, leading to our desired result.

{\bf Proof 2:}~The above proof provides granular details into the convergence to the Uniform distribution on the hypersphere by studying the scaled Chi-distribution. For completeness, we also provide a more straightforward argument, sufficient to study the limiting case. First, it is known that $\frac{Z}{\sqrt{K}}$ being isotropic Gaussian, the distribution of norms, $\|Z\|_2^2/K$, is a Chi-squared distribution with mean $1$ and variance $2/K$. That is, as $K$ increases as the norms distribution converges to a Dirac at $1$. Lastly, because $\frac{Z}{\sqrt{K}}$ is isotropic, it will be uniformly distribution on the hypersphere after normalization. But as $K$ increases, as the samples are already normalized, hence leading to our result.
\end{proof}

\subsection{Proof of \cref{thm:general_density}}
\label{proof:general_density}

\begin{proof}
First and foremost, recall that the density of the random variable $f(X)$ is given by \cref{eq:general}. Relying on \cref{thm:uniform} which stated that for large $K$, our assumption on the output density reads  $f(\vx) \sim \mathcal{U}(0,1)$, we obtain that
$
    \int_{\{\vu \in \mathbb{R}^{D} | f(\vu)=f(\vx)\}} \frac{p_{X}(\vu)}{\prod_{k=1}^{rank(J_{f}(\vu))}\sigma_k(J_{f}(\vu))} \mathrm{d} \mathcal{H}^{r}(\vu)={\rm cst}
$. Now if $f$ is bijective between ${\rm supp}(p_X)$ and $\mathbb{R}^K$, then it is direct to see that 
$
    p_{X}(x) \propto \prod_{k=1}^{rank(J_{f}(\vx))}\sigma_k(J_{f}(\vx))
$. Now if $f$ is surjective there is no longer a one-to-one mapping between $f$ and $p_X$. Instead, there is ambiguity over each level set of $f$. To see that, recall that we only need to maintain a constant value over the integration on the level set. Hence, $f$ is free to scale up one subset of that level set, and scale down another subset, proportionally to $p_X$ to preserve the integration to a constant.
\end{proof}

\subsection{Proof of \cref{thm:JEPA}}
\label{proof:JEPA}
\begin{proof}
    
The role of the predictor in JEPA training is to allow for additional computation to predict one view's embedding from the other view's embedding. While this provides numerous empirical benefits, e.g., in terms of optimization landscape, it actually does not impact the level-set of the encoder--which is what is needed in \cref{eq:general}.

To understand the above argument, consider the case where the views are obtained from applying a transformation such as masking. We denote by $\mathcal{M}$ the masking random and by ${\rm mask}(\vx)$ the application of one realization of $\mathcal{M}$ onto the input $\vx$. We thus have for the invariance term of sample $\vx_n$ 
\begin{align*}
    \mathbb{E}_{({\rm mask}^{(1)},{\rm mask}^{(2)})\sim (\mathcal{M},\mathcal{M})}\left[{\rm dist}\left({\rm Pred}\left({\rm Enc}\left({\rm mask}^{(1)}(\vx_n)\right)\right),{\rm Enc}\left({\rm mask}^{(2)}(\vx)\right)\right)\right].
\end{align*}
Because the predictor is only applied on one of the two embeddings, it is clear that for the JEPA loss to be minimized, it must also be true that
\begin{align*}
    \mathbb{E}_{{\rm mask}^{(2)}\sim \mathcal{M}}\left[{\rm dist}\left({\rm Pred}\left({\rm Enc}\left({\rm mask}^{(1)}(\vx_n)\right)\right),{\rm Enc}\left({\rm mask}^{(2)}(\vx)\right)\right)\right]=0,
\end{align*}
for any realization of ${\rm mask}^{(1)}$. In other word, the encoder's invariance is over the support of $\mathcal{M}$ no matter if the predictor is identity or nonlinear.
Therefore our result directly follows from the above combined with \cref{eq:general,eq:data}.
\end{proof}

\section{Implementation Details}

 \begin{lstlisting}[caption=JEPA-SCORE implementation in PyTorch. Our empirical ablations demonstrate that JEPA-SCORE is not sensitive to the choice of eps (we pick $1e-6$),label=code]
  from torch.autograd.functional import jacobian

  # model returns a tensor of shape (num_samples, features_dim)
  J = jacobian(lambda x: model(x).sum(0), inputs=images)
  with torch.inference_mode():
    J = J.flatten(2).permute(1, 0, 2)
    svdvals = torch.linalg.svdvals(J)
    jepa_score = svdvals.clip_(eps).log_().sum(1)
  \end{lstlisting}

\section{Additional Figures}

\begin{figure}[t!]
    \centering
    \includegraphics[width=\linewidth]{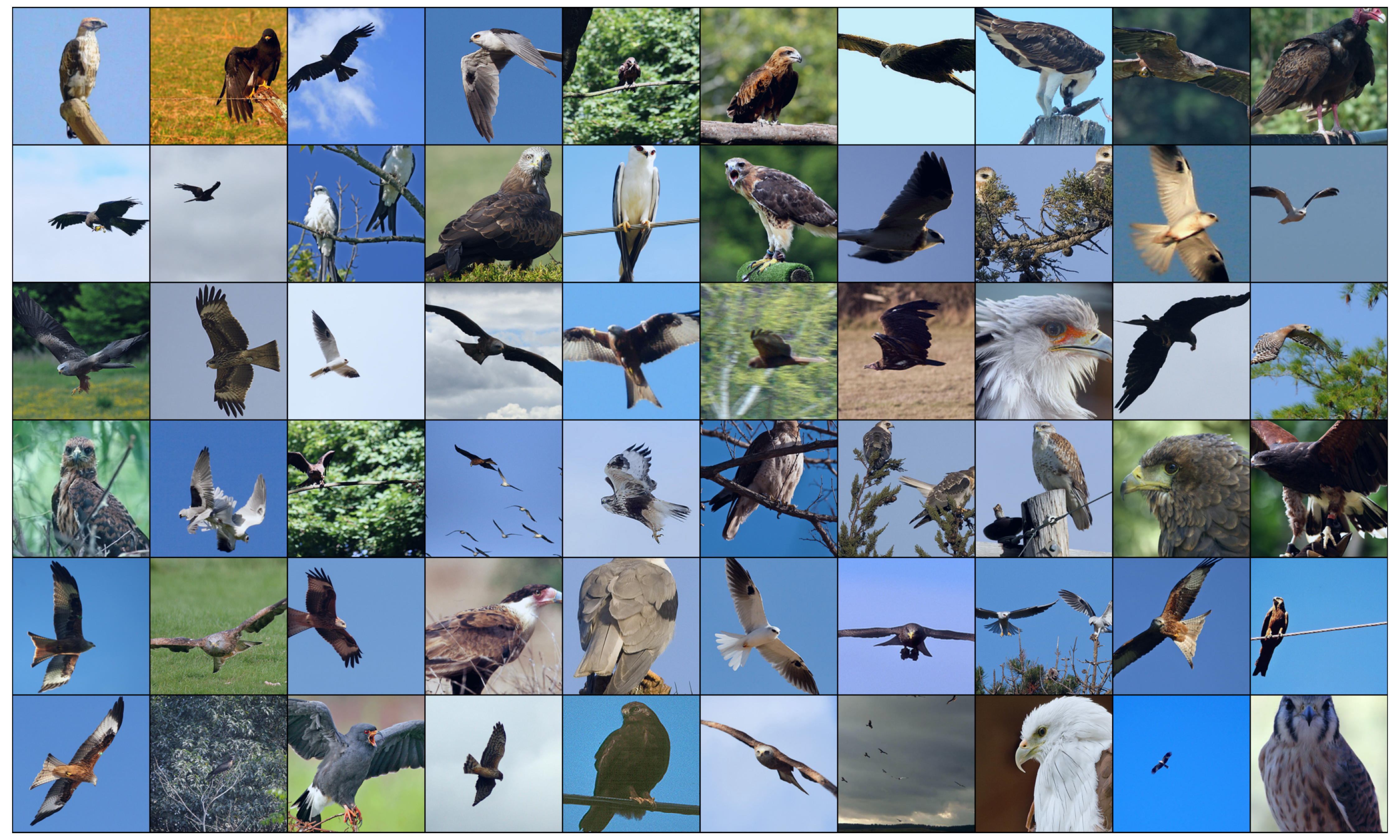}
    \caption{Random samples from Imagenet-1k training dataset for class 21.}
    \label{fig:samples_1}
\end{figure}

\begin{figure}[t!]
    \centering
    \centering
    \begin{minipage}{0.02\linewidth}
    \rotatebox{90}{\small MetaCLIP\;\;IJEPA-22k\;\;IJEPA-1k\;\;DINOv2\;\;\;\;\;\;}
    \end{minipage}
    \begin{minipage}{0.97\linewidth}
    \centering
    \begin{minipage}{0.49\linewidth}
    \centering\color{blue}
        {\bf low probability}\\[-0.6em]
        \makebox[0pt][c]{}\rule{\linewidth}{1.6pt}
    \end{minipage}
    \begin{minipage}{0.49\linewidth}
    \centering\color{red}
    {\bf high probability}\\[-0.6em]
        \makebox[0pt][c]{}\rule{\linewidth}{1.6pt}
    \end{minipage}
    \includegraphics[width=\linewidth]{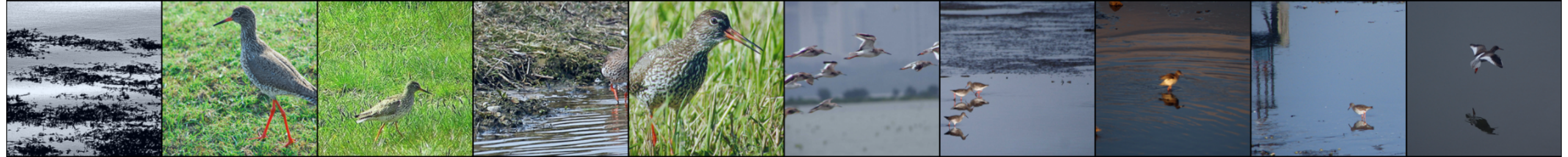}\\
    \includegraphics[width=\linewidth]{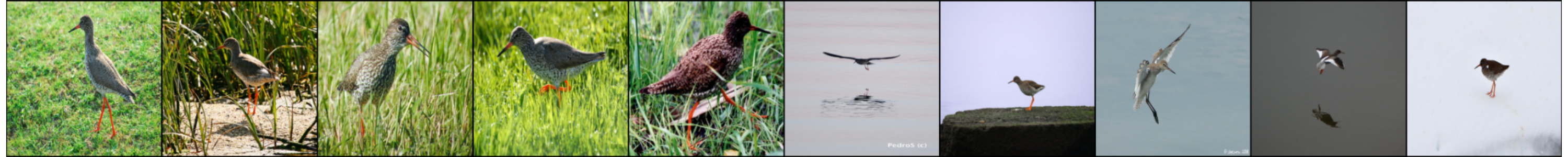}\\
    \includegraphics[width=\linewidth]{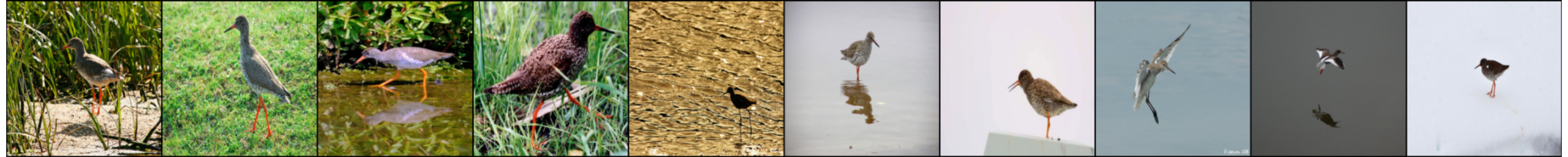}\\
    \includegraphics[width=\linewidth]{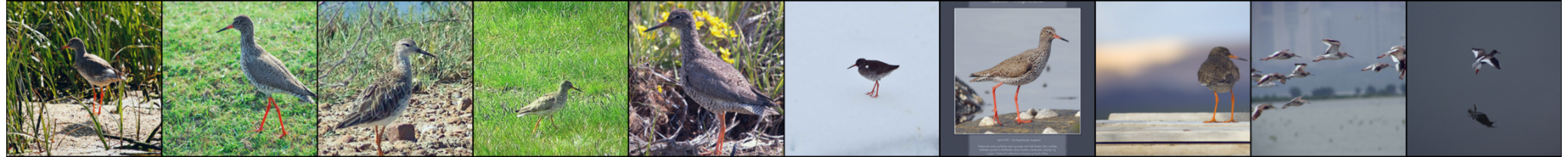}
    \end{minipage}\\
    Random samples\\
    \includegraphics[width=\linewidth]{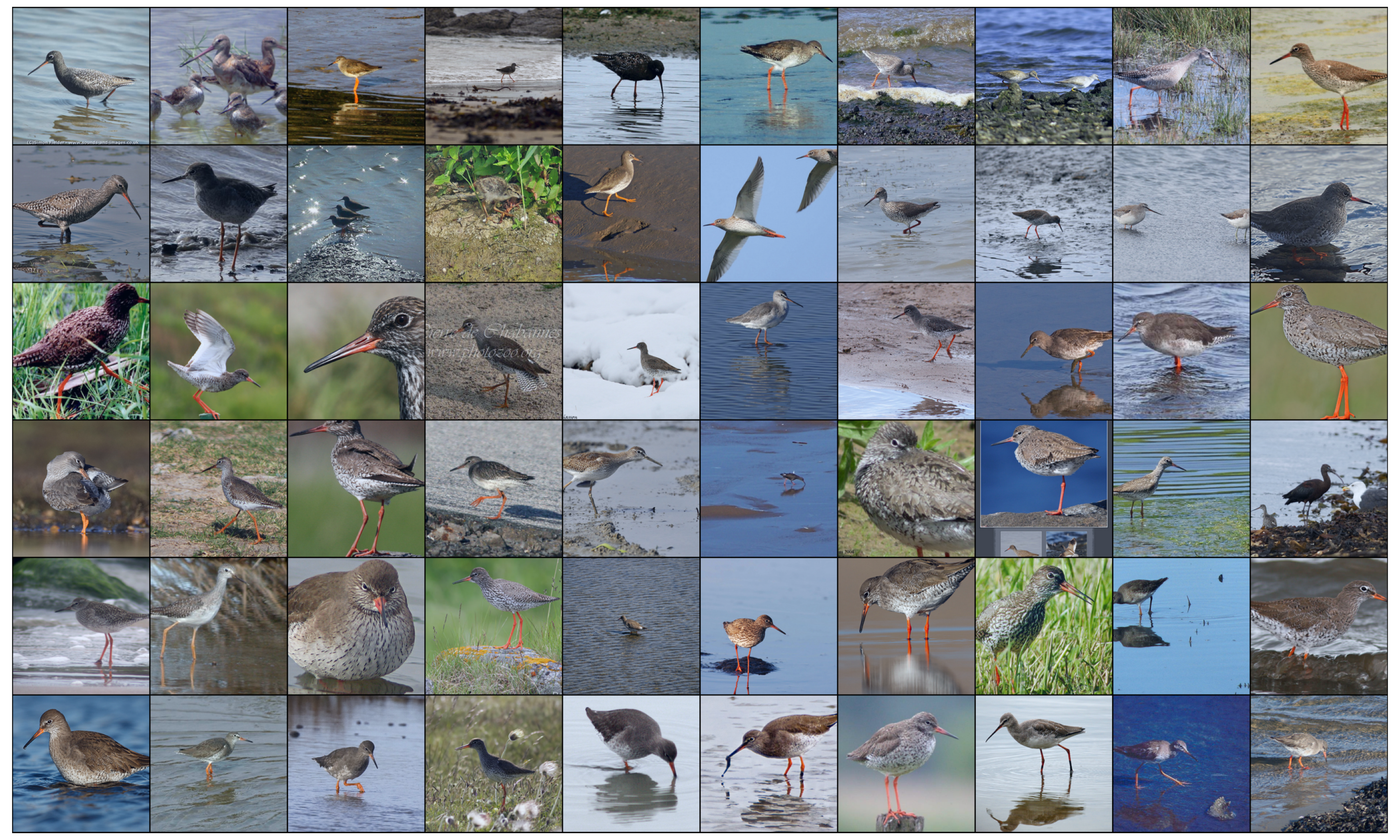}
    \caption{{\bf Top:} least and most likely samples based on the proposed JEPA-SCORE for a fe different pretrained backbones on class 141. {\bf Bottom:}Random samples from Imagenet-1k training dataset for class 141.}
    \label{fig:samples_2}
\end{figure}

\begin{figure}[t!]
    \centering
    \centering
    \begin{minipage}{0.02\linewidth}
    \rotatebox{90}{\small MetaCLIP\;\;IJEPA-22k\;\;IJEPA-1k\;\;DINOv2\;\;\;\;\;\;}
    \end{minipage}
    \begin{minipage}{0.97\linewidth}
    \centering
    \begin{minipage}{0.49\linewidth}
    \centering\color{blue}
        {\bf low probability}\\[-0.6em]
        \makebox[0pt][c]{}\rule{\linewidth}{1.6pt}
    \end{minipage}
    \begin{minipage}{0.49\linewidth}
    \centering\color{red}
    {\bf high probability}\\[-0.6em]
        \makebox[0pt][c]{}\rule{\linewidth}{1.6pt}
    \end{minipage}
    \includegraphics[width=\linewidth]{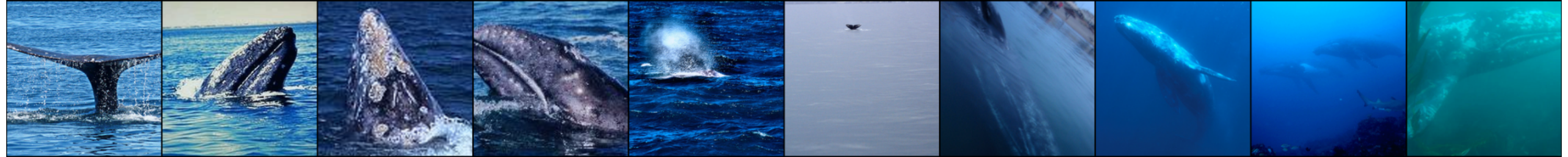}\\
    \includegraphics[width=\linewidth]{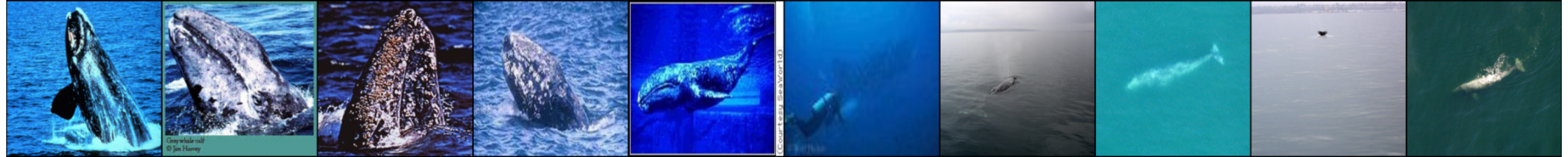}\\
    \includegraphics[width=\linewidth]{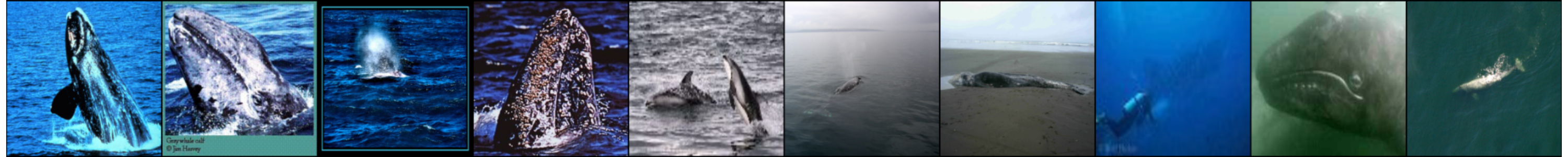}\\
    \includegraphics[width=\linewidth]{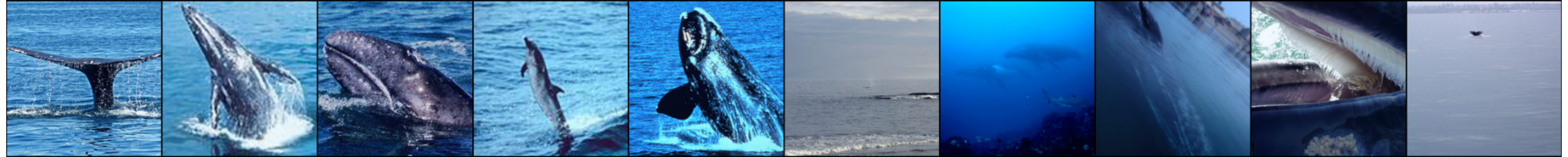}
    \end{minipage}\\
    Random samples\\
    \includegraphics[width=\linewidth]{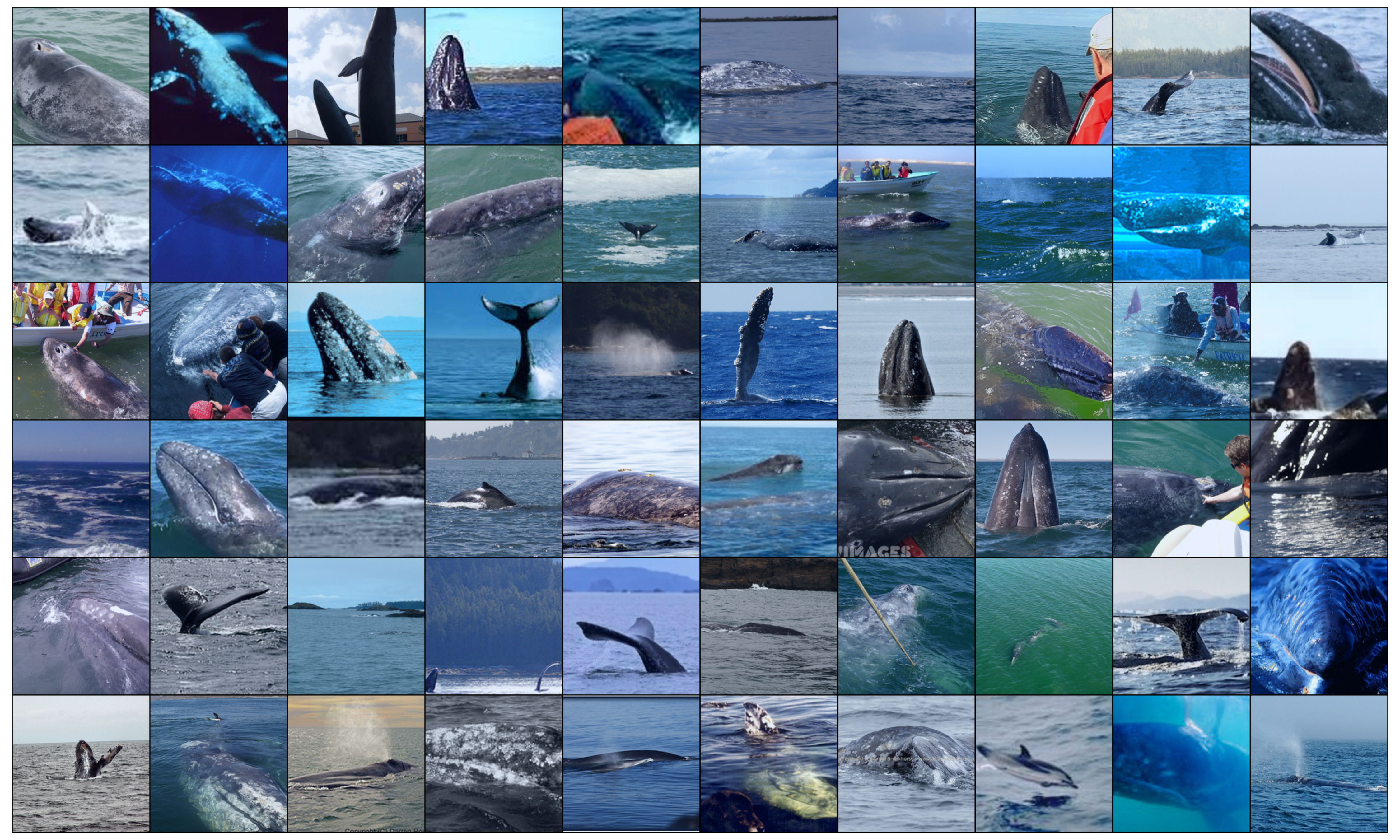}
    \caption{{\bf Top:} least and most likely samples based on the proposed JEPA-SCORE for a fe different pretrained backbones on class 141. {\bf Bottom:}Random samples from Imagenet-1k training dataset for class 147.}
    \label{fig:samples_3}
\end{figure}

\begin{figure}[t!]
    \centering
    \includegraphics[width=\linewidth]{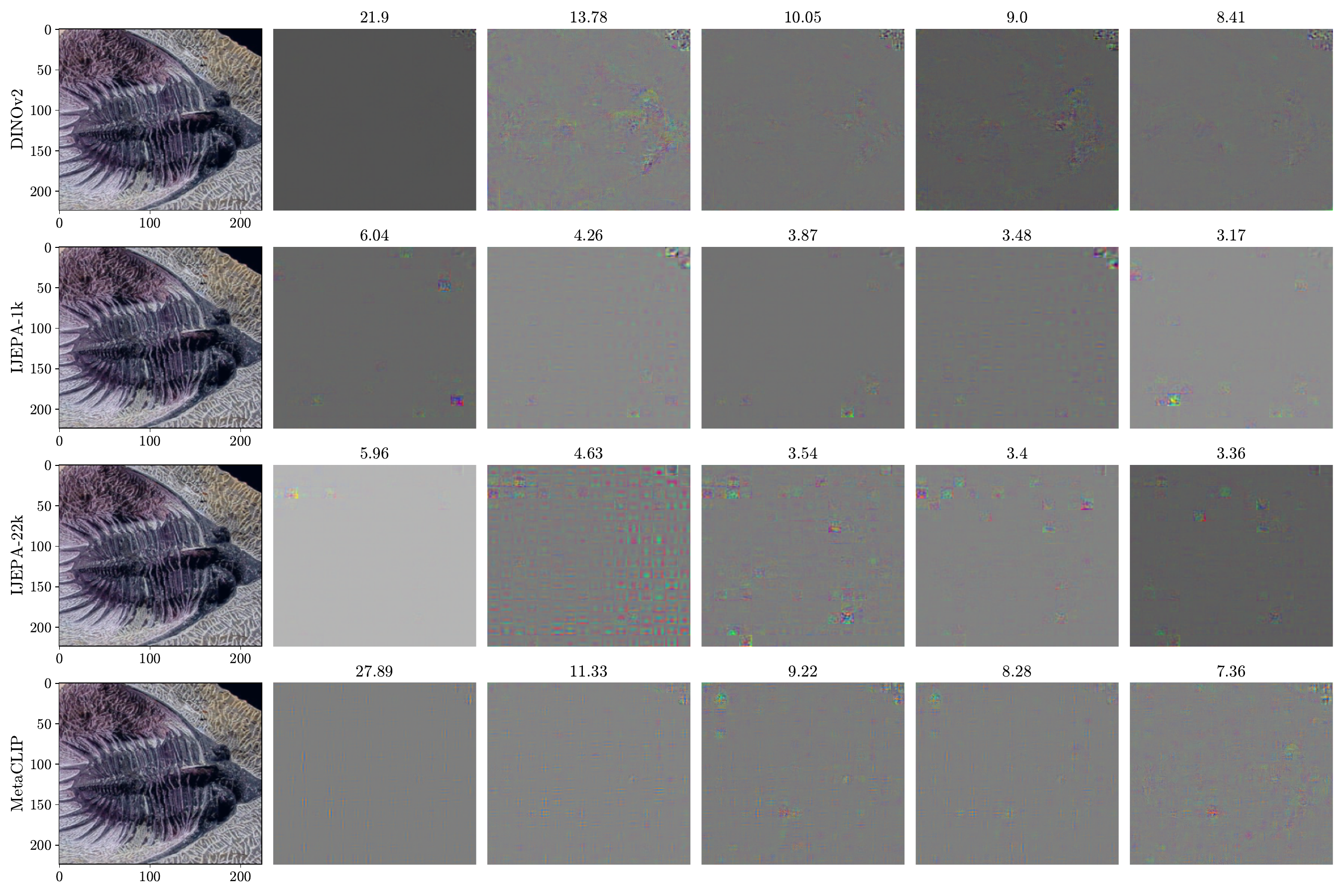}
    \includegraphics[width=\linewidth]{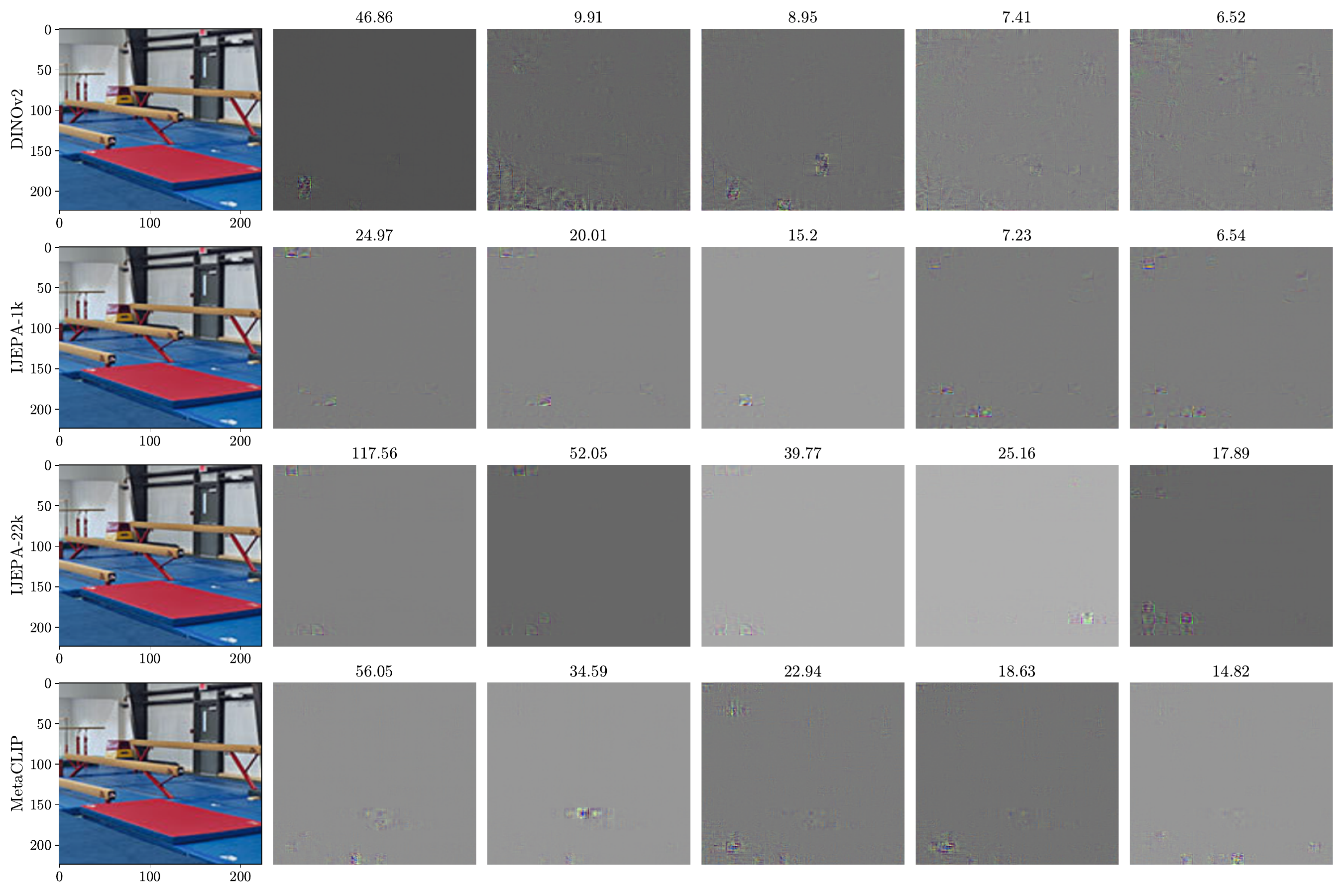}
    \caption{Depiction of the top singular vectors of the Jacobian matrix (columns) for a given input image (left column). The corresponding singular value is provided in the title of each subplot.}
    \label{fig:singular_vectors}
\end{figure}

\newpage

\end{document}